\documentclass[twoside,11pt]{article}

\usepackage{jmlr2e}
\usepackage{amssymb}
\usepackage{mathtools}
\usepackage{comment}
\usepackage{microtype}
\usepackage{booktabs} 
\usepackage{amsmath}

\usepackage{hyperref}

\usepackage{graphicx}
\usepackage{caption} 
\usepackage{url}
\usepackage{multirow}
\usepackage{subcaption}
\usepackage{wrapfig}
\usepackage{lipsum}
\usepackage[ruled]{algorithm2e}

\newcommand{\model}{\textsc{MiLEt}}
\newcommand{\argmax}{\text{argmax}}
\newcommand{\Require}{\textbf{Input:}\,}

\ShortHeadings{Meta-Reinforcement Learning via Exploratory Task Clustering}{Chu and Wang}
\firstpageno{1}

\begin{document}

\title{Meta-Reinforcement Learning via Exploratory Task Clustering}

\author{\name Zhendong Chu \email zc9uy@virginia.edu \\
       \addr Department of Computer Science\\
       University of Virginia\\
       Charlottesville, VA 22903, USA
       \AND
       \name Hongning Wang \email hw5x@virginia.edu \\
       \addr Department of Computer Science\\
       University of Virginia\\
       Charlottesville, VA 22903, USA}

\editor{}

\maketitle

\begin{abstract}
Meta-reinforcement learning (meta-RL) aims to quickly solve new tasks by leveraging knowledge from prior tasks. However, previous studies often assume a single mode homogeneous task distribution, ignoring possible structured heterogeneity among tasks. Leveraging such structures can better facilitate knowledge sharing among related tasks and thus improve sample efficiency. In this paper, we explore the structured heterogeneity among tasks via clustering to improve meta-RL. We develop a dedicated exploratory policy to discover task structures via divide-and-conquer. The knowledge of the identified clusters helps to narrow the search space of task-specific information, leading to more sample efficient policy adaptation. Experiments on various MuJoCo tasks showed the proposed method can unravel cluster structures effectively in both rewards and state dynamics, proving strong advantages against a set of state-of-the-art baselines.
\end{abstract}

\section{Introduction}
Conventional reinforcement learning (RL) is notorious for its high sample complexity, which often requires millions of interactions with the environment to learn a performing policy for a new task. Inspired by the learning process of humans, meta-reinforcement learning (meta-RL) is proposed to quickly learn new tasks by leveraging knowledge shared by related tasks \citep{finn2017model,duan2016rl,wang2016learning}. Extensive efforts have been put into modeling transferable knowledge in meta-RL. For example, \citet{finn2017model} proposed to learn a set of shared meta parameters which are used to initialize the local policy when a new task arrives. \citet{duan2016rl} and \citet{wang2016learning} trained an RNN encoder to characterize prior tasks according to the interaction history in those tasks. 

But little attention has been paid to the structures in the task distribution. All the aforementioned methods implicitly assume tasks follow a uni-modal distribution, and thus knowledge can be broadly shared across all tasks. However, heterogeneity among tasks is not rare in practice, providing subtle structures to better identify transferable knowledge within groups of similar tasks. For instance, the general skills required for the Go game and Gomoku game are related, such as familiarity with the board layout and stone colors. But to achieve mastery in a specific game, policies must acquire and internalize game-specific knowledge/rules in order to effectively navigate subsequent matches. For example, experience about competing against different human players in Go games can be shared within, but not over to Gomoku games. 
This heterogeneity motivates us to formulate a more delicate but also more realistic meta-RL setting where tasks are originated from various but a finite number of distributions, i.e., tasks are clustered. Hence, some knowledge can only be shared within clusters, and it is crucial to leverage this information to improve sample efficiency of task modeling. We refer to this as \emph{structured heterogeneity} among RL tasks, and explicitly model it in the task distribution to capture cluster-level knowledge\footnote{We do not assume the knowledge in different clusters is exclusive, and thus each cluster can still contain overlapping knowledge, e.g., motor skills in locomotion tasks.}.

Structured heterogeneity among tasks has been studied in supervised meta-learning \citep{yao2019hierarchically}; but it is a lot more challenging to be handled in meta-RL, where the key bottleneck is \emph{how to efficiently discover the clustering structure in a population of RL tasks.} Different from supervised learning tasks where static task-specific data are available before any learning starts, the observations about RL tasks are collected by an agent's interactions with the task environment. More specifically, successfully adapting a RL policy to a new task depends on accurate profiling of the task, which however is elicited by the policy itself, i.e., the chicken-and-egg dilemma \citep{liu2021decoupling}. 
To break the circle, we propose to a divide-and-conquer exploration strategy to build accurate task profiles. Our approach explores at the cluster level, which is a broader abstraction of similar tasks and is expected to be identified with fewer samples in a new task. With the identified task, the agent can focus more on exploring task-specific information within a narrowed search space, resulting in improved sample efficiency.

To realize our idea of discovering structured heterogeneity of tasks in meta-RL, we propose a cluster-based meta-RL algorithm, called \model: \textbf{M}eta re\textbf{I}nforcement \textbf{L}earning via \textbf{E}xploratory Task clus\textbf{T}ering, which is designed to explore clustering structures of tasks and achieve fast adaptation in new tasks via cluster-level transferable knowledge sharing. To the best of our knowledge, we are the first to propose a method for improving sample efficiency in meta-RL by utilizing cluster structures in the task distribution. Specifically,
we perform cluster-based posterior inference \citep{rao2019continual, dilokthanakul2016deep} to infer the cluster of a new task according to its ongoing trajectory. To facilitate cluster inference in a new task, at the meta-train phase, we optimize a dedicated exploration policy designed to reduce uncertainty in cluster inference as the agent's interaction with the task environment progresses. An exploitation policy is then trained to maximize the task rewards within the narrowed search space suggested by the identified cluster. 

We compare \model{} against a rich set of state-of-the-art meta-RL solutions on various MuJoCo environments \citep{todorov2012mujoco} with varying cluster structures in both reward and state dynamics. We also show our method can mitigate the sparse reward issue by sample-efficient exploration on cluster structures. To test the generality of our solution, we test in environments without explicit clustering structure among tasks, and the results show \model{} can still discover locally transferable knowledge among the observed tasks and benefit fast adaptation to related new tasks. 
\section{Related work}
\textbf{Task modeling in meta-learning.} Task modeling is important to realize fast adaptation in new tasks in meta learning. \citet{finn2017model} first proposed the model-agnostic meta learning (MAML) aiming to learn a shared model initialization, i.e., the meta model, given a population of tasks. MAML does not explicitly model tasks, but it expects the meta model to be only a few steps of gradient update away from all tasks. 
Later, an array of methods extend MAML by explicitly modeling tasks using given training data under the supervised meta-learning setting. \citet{lee2018gradient} learned a task-specific subspace of each layer's activation, on which gradient-based adaptation is performed. \citet{vuorio2019multimodal} explicitly learned task embeddings given data points from each task and then used it to generate task-specific meta model. \citet{yao2019hierarchically} adopted a hierarchical task clustering structure, which enables cluster-specific meta model. Such a design encourages the solution to capture locally transferable knowledge inside each cluster, similar to our \model{} model. 
However, task information is not explicitly available in meta-RL: since the true reward/state transition functions are not accessible to the agent, the agent needs to interact with the environment to collect observations about the tasks, while maximizing the return from the interactions. \model{} performs posterior inference of a task's cluster assignment based on its ongoing trajectory; better yet, it is designed to behave exploratorily to quickly identify tasks' clustering structures, and then refine the task modeling in the narrowed search space conditional on the identified cluster. 
 
\textbf{Exploration in meta-reinforcement learning.} Exploration plays an important role in meta-RL, as the agent can only learn from its interactions with the environment. In gradient-based meta-RL \citep{finn2017model}, the local policy is trained on the trajectories collected by the meta policy, and thus the exploration for task structure is not explicitly handled. \citet{Stadie2018SomeCO} and \citet{rothfuss2018promp} computed gradients w.r.t. the sampling distribution of the meta policy, in addition to the collected trajectories. \citet{gupta2018meta} also extended MAML by using learnable latent variables to control different exploration behaviors.  The context-based meta-RL algorithms \citep{duan2016rl, wang2016learning} automatically learn to trade off exploration and exploitation by learning a policy conditioned on the current context. \citet{Zintgraf2020VariBAD:} explicitly provided the task uncertainty to the policy to facilitate exploration. \citet{zhang2021metacure} and \citet{liu2021decoupling} developed a separate exploration policy by maximizing the mutual information between task ids and inferred task embeddings.  However, all the aforementioned methods did not explicitly explore at the cluster level, which will miss the opportunity to efficiently obtain task-specific information when explicit clustering structures exist in the task distribution. \model{} first explores to identify the cluster of a task, which is expected to take less samples than what is needed to identify the detailed tasks; then the agent can explore task information within a narrower search space for improved sample efficiency.

\section{Background}
\noindent\textbf{Meta-reinforcement learning.} We consider a family of Markov decision processes (MDPs) \footnote{The terms of environment, task and MDP are used interchangeably in this paper, when no ambiguity is incurred.} $p(\mathcal{M})$, where an MDP $M_i\sim p(\mathcal{M})$ is defined by a tuple $M_i=(\mathcal{S}, \mathcal{A}, R_i, T_i, T_{i, 0}, \gamma, H)$ with $\mathcal{S}$ denoting its state space, $\mathcal{A}$ as its action space, $R_i(r_{t+1}|s_t, a_t)$ as its reward function, $T_i(s_{t+1}|s_t, a_t)$ as its state transition function, $T_{i,0}(s_0)$ as its initial state distribution, $\gamma$ as a discount factor, and $H$ as the length of an episode. The index $i$ represents the task id, which is provided to agents in some works \citep{zhang2021metacure, liu2021decoupling, rakelly2019efficient}.
We consider a more general setting where the task id is not provided to the agent \citep{Zintgraf2020VariBAD:}, as in general we should not expect the task id would encode any task-related information. Tasks sampled from $p(\mathcal{M})$ typically differ in the reward and/or transition functions. In each task, we run a \emph{trial} consisting of $N+1$ episodes \citep{duan2016rl}. Following the evaluation settings in previous works \citep{finn2017model, liu2021decoupling, zhang2021metacure, rothfuss2018promp}, the first episode in a trial is reserved as an \emph{exploration} episode to gather task-related information, and an agent is evaluated by the returns in the following $N$ \emph{exploitation} episodes. 

Inside a trial, we denote the agent's interaction with the MDP at time step $n$ as $\tau_n=\{s_n, a_n, r_n, s_{n+1}\}$, and $\tau_{:t}=\{s_0, a_0, r_0, ..., s_t\}$ denotes the interaction history collected before time $t$. In the exploration episode, an agent should form the most informative trajectory $\tau^\psi$ by rolling out an exploration policy $\pi_\psi$ parameterized by $\psi$. In exploitation episodes, the agent executes the exploitation policy $\pi_\phi$ parameterized by $\phi$ (in some prior work, $\pi_\psi$ and $\pi_\phi$ are the same) conditioned on $\tau^\psi$ and, optionally, the history collected in exploitation episodes $\tau^\phi$. The returns in exploitation episodes is computed as,
\begin{equation}
\!\!\!\mathcal{J}(\pi_\psi, \pi_\phi) = \mathbb{E}_{M_i \sim p(\mathcal{M}), \tau^\psi\sim \pi_\psi}\Big [\sum_{t=0}^{N\times H}R_i\big (\pi_\phi(\tau^\psi; \tau^\phi_{:t}) \big )\Big ], 
    \label{eq:expected_reward}
\end{equation}
where $R_i\big (\pi_\phi(\tau^\psi; \tau^\phi_{:t}) \big )$ is the return of $\pi_\phi$ conditioned on $\tau^\psi$ and $\tau^\phi_{:t}$ at time step $t$ in task $M_i$. 

\noindent\textbf{Clustered RL tasks.}  In this paper, we consider a more general and realistic setting, where the task distribution form a mixture, 
\begin{equation}
    p(\mathcal{M}) = \sum_{c=1}^C w_c \cdot p_c(\mathcal{M}),
    \label{eq:task-mixture}
\end{equation}
where $C$ is the number of mixing components (i.e., clusters) and $w_c$ is the corresponding weight of component $c$, such that $\sum_{c=1}^C w_c=1$. Every task is sampled as follows,
\begin{enumerate}
    \item Sample a cluster $c$ according to the multinomial distribution of $Mul(w_1, ..., w_C)$;
    \item Sample a reward function $R$ or a transition function $T$ or both from $p_c(\mathcal{M})$.
\end{enumerate}
The knowledge shared in different clusters could be different. For example, two clusters of target positions can exist in a navigational environment and each cluster centers on a distinct target position, e.g., top-left vs., bottom-right. 
The knowledge about how an agent reaches the top-left target positions in the first cluster cannot help tasks in the second cluster; but it is crucial for different tasks in the first cluster. In this example, when handling a new task, a good exploration strategy should first explore to identify the cluster of the task (i.e., to move top-left or bottom-right), and then identify the specific target position in the corresponding region of the map. With the identified cluster, the search space can be largely reduced, allowing for more efficient exploration of task-specific information.

\section{Methodology}
\begin{figure}[t]
    \centering
    \includegraphics[width=12cm]{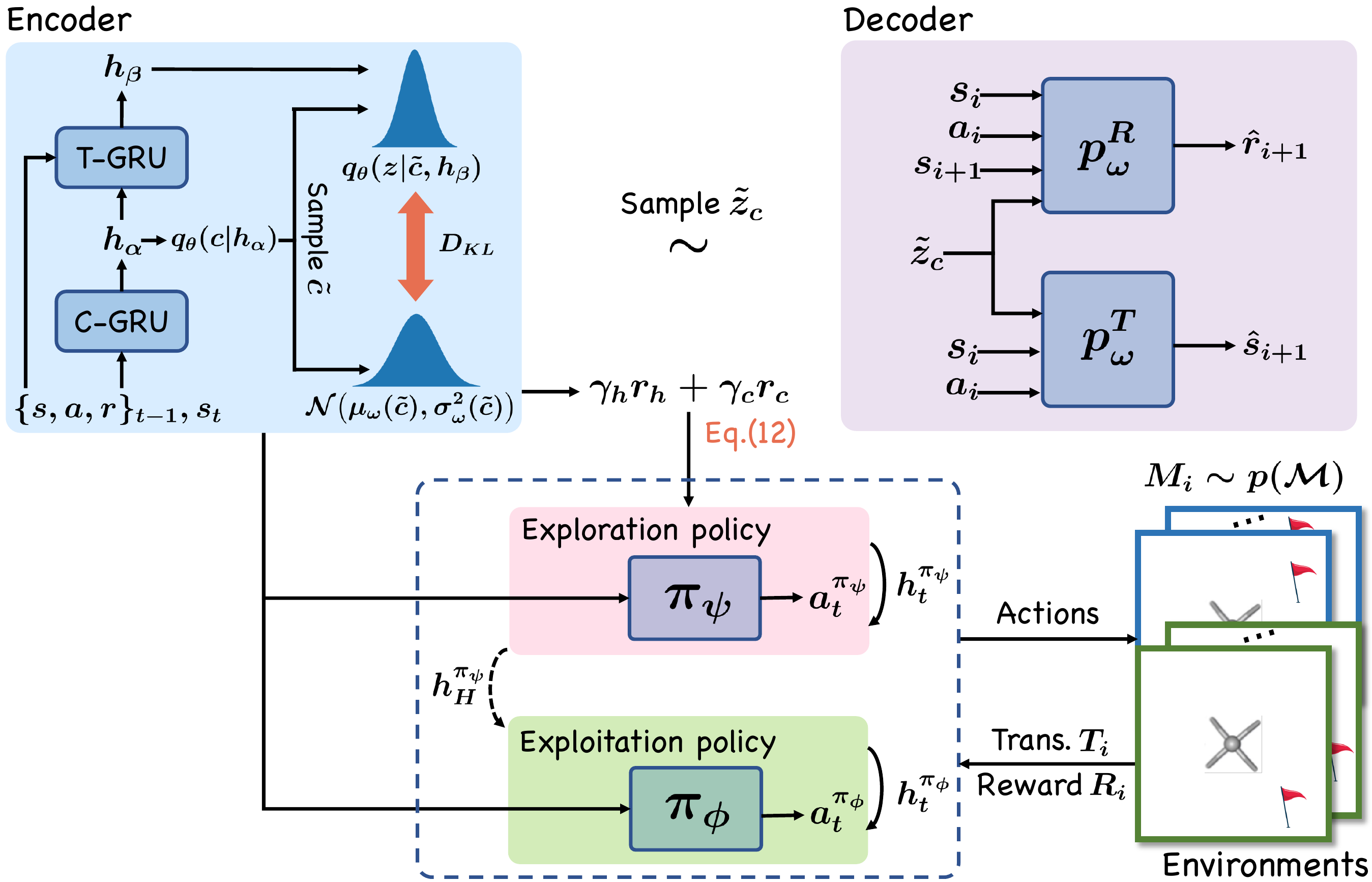}
    \caption{\model{} architecture. The encoder processes ongoing trajectories and performs CBVI for $q_\theta(z|c, h_\beta)$. The exploration policy $\pi_\psi$ is trained to find the most certain cluster assignment $c$ when interacting with the environment. The explored information is passed to the exploitation policy $\pi_\phi$ to facilitate fast adaptation in the inferred task $M_{z_c}$.}
    \label{fig:framework}
    \vspace{-6mm}
\end{figure}

In this section, we present \model{}  in detail. First, we introduce how to estimate population-level task structures using the collected trajectories via cluster-based variational inference (CBVI). Then, we explain the exploration policy trained by the exploration-driven reward, which is designed to quickly identify the cluster assignment of a new task. At a high level, in each task \model{} first executes the exploration policy to collect the coarse-grained cluster information; then it adapts the task policy under the help of the inferred posterior cluster distribution. The architecture of \model{} is shown in Figure \ref{fig:framework}.

\subsection{Cluster-based Variational Inference with Consistency Regularization}
\label{sec:inference}

Since the reward and transition functions are unknown to the agent, we estimate a latent random variable $c_i$ to infer the cluster assignment of current task $M_i \sim p_c(\mathcal{M})$. Based on $c_i$, we infer another latent random variable $z_i$ carrying task-level information, i.e., $z_i$ suggests the reward/transition functions that define the task. For simplicity, we first drop the script $i$ in this section, as we will only use one task as an example to illustrate our model design. 

In meta-RL, all information about a given task can be encoded by $z$. But inferring $z$ can be sample inefficient, as the task space can be very large.  Thanks to the structured heterogeneity among tasks, inferring a task's cluster assignment $c$ can be more sample efficient, since we should expect a much smaller number of task clusters than the number of tasks. Once $c$ is identified, $z$ can be more efficiently identified, i.e., divide and conquer. Hence, in \model, when a new task arrives, we decode its characteristics by the posterior distribution $p(z, c | \tau_{:t})=p(z | \tau_{:t}, c)p(c | \tau_{:t})$ with respect to the interaction history up to time $t$. The inferred task information $z_c$, which refers to $z$ conditioned on $c$, is then provided to the policy $\pi_{\psi/\phi}(a_t|s_t, z_c)$.

However, the exact posterior $p(z, c|\tau_{:t})$ defined by Eq.(\ref{eq:task-mixture}) is intractable. Instead, we learn an approximated variational posterior $q_\theta(z, c|\tau_{:t})=q_\theta(z|\tau_{:t}, c)q_\theta(c|\tau_{:t})$, in which we estimate two dependent inference networks and collectively denote their parameters as $\theta$. On top of the inference networks, we learn a decoder $p_\omega$ to reconstruct the collected trajectories. The whole framework is trained by maximizing the following objective, 
\begin{equation}
    \mathbb{E}_{\rho_\pi(\mathcal{M}, \tau^+)}\Big[\log p(\tau^+|\pi)\Big],
    \label{eq:likelihood}
\end{equation}
where $\rho_\pi$ is the distribution of trajectories induced by the policies $\pi=\{\pi_\psi, \pi_\phi\}$ within the task, and $\tau^+=\{\tau^\psi, \tau^\phi\}$ denotes all trajectories collected in a trial, the length of which is denoted as $H^+=(N+1)H$.  
We choose to use trajectories from both exploration and exploitation episodes to best utilize information about the same underlying MDP. We omit the dependencies on $\pi$ to simplify our notations in later discussions. Instead of optimizing the intractable objective in Eq.(\ref{eq:likelihood}), we optimize its evidence lower bound (ELBO) w.r.t. the approximated posterior $q_\theta(z, c|\tau_{:t})$ estimated via Monte Carlo sampling \citep{rao2019continual} (full derivation can be found in Appendix \ref{app:elbo}),
\begin{align}
       ELBO_t = ~\mathbb{E}_\rho \biggr[& \overset{\text{cluster-specific reconstruction likelihood}}{\overbrace{\mathbb{E}_{q_\theta(z, c|\tau_{:t})} \big[\ln p_\omega(\tau^+|\Tilde{z}_c) \big ]}} \nonumber \\
       & - \overset{\text{cluster-specific regularization}} {\overbrace{\mathbb{E}_{q_\theta(c|\tau_{:t})} \big[ \text{KL}(q_\theta(z|c, \tau_{:t}) \parallel p_\omega(z|c)) \big]}} \nonumber \\
   & - \overset{\text{cluster regularization}}{\overbrace{\text{KL}(q_\theta(c|\tau_{:t})\parallel p(c))}}\biggr],
\label{eq:elbo}
\end{align}
where $p_\omega(z|c) = \mathcal{N}\big(\mu_\omega(c), \sigma^2_\omega(c)\big)$ is a learnable cluster-specific prior, which is different from the simple Gaussian prior used in single-mode VAE \citep{kingma2013auto}.  This prior allows \model{} to capture unique characteristics of each cluster. $p_\omega(z|c)$'s parameters are included in $\omega$ since the cluster structure is also a part of the environment. $\Tilde{z}_c$ is the latent variable sampled from $q_\theta(z|c, \tau_{:t})=\mathcal{N}\big(\mu_\theta(c, \tau_{:t}), \sigma^2_\theta(c, \tau_{:t})\big)$, using the reparameterization trick \citep{kingma2013auto}. $q_\theta(c| \tau_{:t})$ outputs the approximated posterior cluster distribution given $\tau_{:t}$\footnote{In practice, we use the Gumbel-softmax trick to simplify the calculation.}. $p(c)$ is a fixed non-informative multinomial distribution representing the prior cluster distribution in tasks. Intuitively, 
if discrete structures (i.e., clusters) exist in the task distribution, a uniform $q_\theta(c|\tau_{:t})$ (i.e., collapsed) will cause low reconstruction likelihood. As a result, clustering is preferred when specific cluster assignments can reconstruct the trajectories well.

Similar to \cite{Zintgraf2020VariBAD:}, the first term $\ln p_\omega(\tau^+|\Tilde{z}_c)$ in Eq.(\ref{eq:elbo}) can be further factorized as,
\begin{align*}
        \ln p_\omega(\tau^+|\Tilde{z}_c, \pi) =& \ln p(s_0|\Tilde{z}_c)  + \sum_{i=0}^{H^+-1}\Big[\ln p_\omega(s_{i+1}|s_i, a_i, \Tilde{z}_c)  + \ln p_\omega(r_{i+1}|s_i, a_i, s_{i+1}, \Tilde{z}_c)\Big] \\
          \approx & \, \text{const.} + \sum_{i=0}^{H^+-1} -\big(r_i - \hat{r}(s_i, a_i, s_{i+1}, \Tilde{z}_c)\big)^2  - \lambda_s \left\|s_{i+1}
          - \hat{s}(s_i, a_i, \Tilde{z}_c) \right\|_2^2,
\end{align*}
where $p(s_0|\Tilde{z}_c)$ is the initial state distribution in a task, and we consider it as a constant by assuming identical distribution of the initial states across clusters. The second and third terms are likelihood derived from the decoders for transition and reward functions. $\lambda_s$ control the approximation of the state transition in variational inference.
The density functions of $p_\omega(s_{i+1}|s_i, a_i, \Tilde{z}_c)$ and $p_\omega(r_{i+1}|s_i, a_i, s_{i+1}, \Tilde{z}_c)$ are difficult to estimate in continuous state and action spaces. Denote the corresponding decoder output as $\hat{r}$ and $\hat{s}$, we use L2 distance to approximate the log-likelihood functions \citep{zhang2021metacure}.


In the inference networks $q_\theta(z|\tau_{:t}, c)$ and $q_\theta(c|\tau_{:t})$, we follow \cite{duan2016rl, Zintgraf2020VariBAD:} to encode the history $\tau_{:t}$ by Gated Recurrent Units (GRUs) \citep{chuang2014gru}. We propose a stacked GRU structure (shown in Figure \ref{fig:framework}) to differentiate the information for cluster and task inference in the hidden space. Specifically, we set a task-GRU (T-GRU) and a cluster-GRU (C-GRU), both of which encode the history $\tau_{:t}$, but with different levels of granularity. T-GRU is set to capture fine-grained task-specific patterns in the history, as it is optimized to reconstruct trajectories of a specific task. C-GRU captures more coarse-grained patterns beyond tasks, as it is set to help T-GRU reconstruct all trajectories within a cluster. To realize this difference, the output $h_\beta$ of T-GRU is only provided to $q_\theta\big(z|h_\beta(\tau_{:t}, h_\alpha), c\big)$, while the output $h_\alpha$ of C-GRU is passed to both cluster inference $q_\theta\big(c|h_\alpha(\tau_{:t})\big)$ and task inference $q_\theta\big(z|h_\beta(\tau_{:t}, h_\alpha), c\big)$. 
This also reflects our dependency assumption about the task structure: cluster assignment determines tasks. We denote $h=\{h_\alpha, h_\beta\}$, and $h$ is passed across episodes in a trial.

Different from the static training data provided beforehand in supervised meta-learning settings, the trajectory data is incrementally collected by the agent in meta-RL, which brings both challenges and opportunities in obtaining the most informative information for cluster inference. 
First, inside a trial, the inference improves as more observations are collected, which means the agent's belief about the ongoing task could change thereby. 
This is problematic, since the cluster inference result should stay consistent within a given task, no matter how trajectory changes over episodes. We attribute this property as \textit{in-trial} consistency. It can be measured by $\text{KL}(q(c|\tau_{:t_1})\parallel q(c|\tau_{:t_2}))$, where $t_1$ and $t_2$ refer to two arbitrary timestamps in a trial.
We enforce the notion of cluster inference consistency via the following regularizer,
\begin{equation}
    \mathcal{L_{\text{I}}} = \frac{1}{H^+-1}\sum_{t=0}^{H^+-1}\text{KL}\big(q_\theta(c|\tau_{:t})\parallel q_\theta(c|\tau_{:t+1})\big),
\end{equation}
which minimizes the inconsistency between any two consecutive time steps inside a trial. 

Similarly, since the cluster-specific prior $p_\omega(z|c)$ is learnable, the task inference can become inconsistent if the prior changes drastically across training epochs. 
More seriously, oscillation in the inference of latent variable $z$ can cause the collapse of policy training, as tasks across clusters might be assigned with the same latent variable $z$ across different training epochs. 
We conclude it as the \emph{prior} consistency property and enforce it via the following regularization, 
\begin{equation}
    \mathcal{L}_{\text{P}} = \frac{1}{C}\sum_{c=1}^C\text{KL}\big(p_\omega(z|c) \parallel p_{\text{tgt}}(z|c)\big), 
\end{equation}
where $p_{\text{tgt}}(z|c)$ is a target network and its parameters are the same as $p_\omega(z|c)$ but updated in a much slower pace. We finally obtain the objective in CBVI as follows,
\begin{equation}
    \mathcal{L}(\theta, \omega) = \mathbb{E}_{p(\mathcal{M})}\biggr[\sum_{t=0}^{H+}ELBO_t - \lambda_\text{I}\mathcal{L}_\text{I} - \lambda_\text{P}\mathcal{L}_\text{P} \biggr],
    \label{eq:vae_loss}
\end{equation}
where $\lambda_\text{I}$ and $\lambda_\text{P}$ are hyper-parameters to control the strength of two regularizers.

\subsection{Exploration for Reducing Task Inference Uncertainty}
In \model{}, policy adaptation in a new task has two objectives: (1) explore clustering structure; (2) explore task-specific information to solve the task. 
As we explained before, \model{} follows a divide-and-conquer principle to realize these two objectives, which is implemented by learning two separate policies as shown in Figure \ref{fig:framework}. One takes exploratory behaviors to collect cluster and task information, i.e., the exploration policy $\pi_\psi$. The other is optimized to solve the task with the collected information, i.e., the exploitation policy $\pi_\phi$.

 Clustering structures provide hints to solve specific tasks. We train a dedicated exploration policy to provide a good basis for task-solving by exploring the environment. We evaluate the quality of exploration using two principles. First, whether the trajectory of an exploration episode can reduce the uncertainty of cluster inference. Second, whether the inference result is consistent. 
 We conclude them as \emph{certain} and \emph{consistent} exploration.
  
  We design two intrinsic rewards to encourage certain and consistent inference results. First, we use the entropy of cluster inference network $q_\theta(c|\tau_{:H}^\psi)$ to measure the \emph{uncertainty} of the inferred cluster. For a new task, we look for trajectories that provide the most certain cluster inference. We formalize the objective as follows, omitting the subscript $\theta$ and superscript $\psi$ for simplicity,
\begin{align*}
        H(q(c|\tau_{:H})) = -\mathbb{E}\big[ \ln q(c|\tau_{:H}) \big] = -\mathbb{E}\big[\ln q(c|\tau_0) + \sum_{t=0}^{H-1} \ln \frac{q(c|\tau_{:t+1})}{q(c|\tau_{:t})} \big].
\end{align*}
We then define an intrinsic reward of each action by telescoping the second term similar to \cite{zhang2021metacure, liu2021decoupling},
\begin{align*}
       r_h(a_t) &= \mathbb{E}\big[\ln \frac{q(c|\tau_{:t+1}=[s_{t+1};a_t;r_t;\tau_{:t}])}{q(c|\tau_{:t})} \big] =  H(q(c|\tau_{:t})) - H(q(c|\tau_{:t+1})).
\end{align*}
This reward favorites actions which can reduce the entropy of cluster inference; and therefore, a trajectory leading to a consistent cluster inference is preferred. To more explicitly measure the divergence between the posterior cluster distributions in two steps, we define another reward encouraging consistent cluster inference,
\begin{align*}
    r_c(a_t) &= -\text{KL}(q(c|\tau_{:t})\parallel q(c|\tau_{:t+1})).
\end{align*}
Intuitively, after locating the cluster, the exploration policy should focus on identifying task-level information, which becomes easier within the narrowed search space confined by the identified cluster, i.e., divide-and-conquer. We define the following composed reward to encourage this coarse-to-fine exploration behavior,
 \begin{align}
     r_e(a_t) = r(a_t) + \gamma_h(t)r_h(a_t) + \gamma_c(t)r_c(a_t),
     \label{eq:exploration_reward}
 \end{align}
 where $r(a_t)$ is the reward provided by the environment. $\gamma_h(t)$ and $\gamma_c(t)$ are two temporal decaying functions,
 \begin{align}
         \gamma_h(t) = b_h - a_h\exp(-s_h(H-t)),  \gamma_c(t) = -b_c + a_c\exp(-s_c(H-t)),
 \end{align}
where $\{a, b, s\}_{h, c}$ are hyper-parameters controlling the rate of decay.  $\gamma_h(t)$ should gradually decrease to 0, which encourages the policy to find a certain cluster at the early stage. $\gamma_c(t)$ gradually increases from a negative value to positive. At the early stage, a negative $\gamma_c(t)$ encourages the policy to try different clusters. Later, a positive $\gamma_c(t)$ enforces the policy to stick to the current cluster and focuses more on discovering task information by maximizing raw rewards.

Finally, the exploitation policy $\pi_\phi$ inherits the hidden state $h^{\pi_\psi}_H$, which encodes knowledge collected by the exploration policy, and is then trained to maximize the expected reward defined in Eq.(\ref{eq:expected_reward}). The detailed pseudo-codes of meta-train and meta-test phases for \model{} are shown in Appendix \ref{sec:alg}.

\begin{figure*}[t]
    \begin{subfigure}{1\textwidth}
        \centering
        \includegraphics[width=14.5cm]{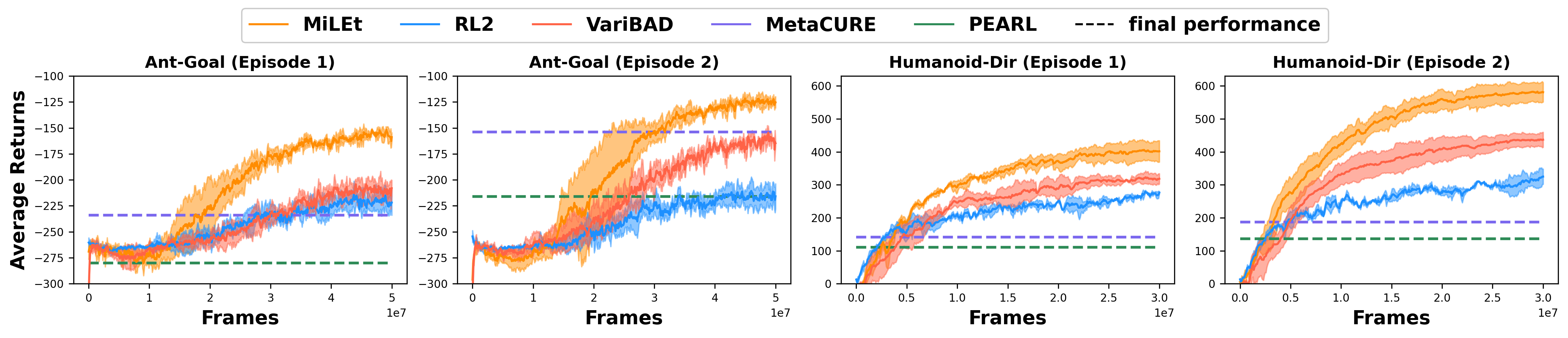}
        \caption{Environments with clustered reward functions.}
        \label{fig:reward_env}
    \end{subfigure}
    ~~
    \begin{subfigure}{1\textwidth}
    \centering
    \includegraphics[width=14.5cm]{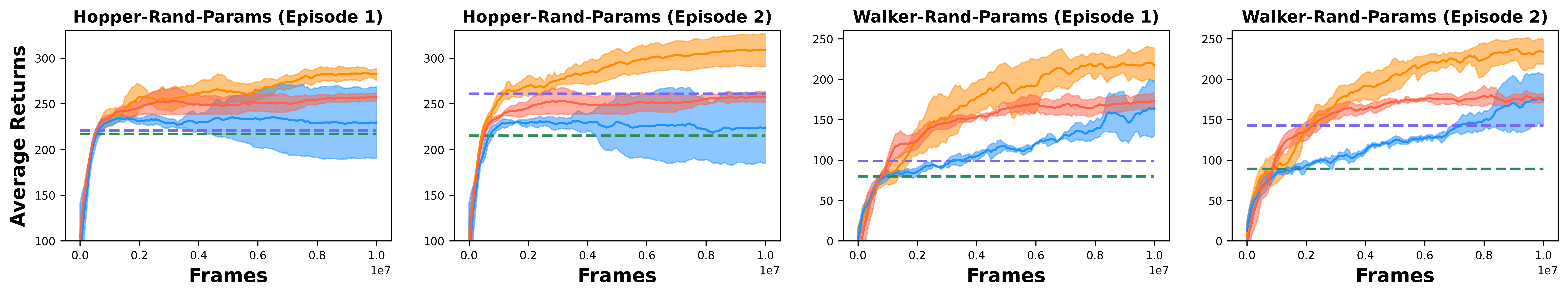}
    \caption{Environments with clustered state transition functions.}
    \label{fig:trans_env}
    \end{subfigure}
    \vspace{-3mm}
    \caption{Average test performance for 2 episodes on MuJoCo environments.}
    \label{fig:cluster_result}
    \vspace{-2mm}
\end{figure*}

\section{Experiments}

In this section, to fully demonstrate the effectiveness of \model{} in handling structured heterogeneity in meta-RL,  we conduct a set of extensive experiments to study the following research questions: (1) Can \model{} achieve better performance than state-of-the-art meta-RL algorithms by exploring structured heterogeneity in the task distribution? (2) Can \model{} effectively discover cluster structures in both rewards and state dynamics? (3) Can the sparse reward issue be mitigated by exploratory clustering of tasks? (4) How does the number of clusters affect the final performance of \model{}? (5) Can \model{} be capable of clustering entire tasks based on the more abstract problem being solved, e.g., the Meta-World task suite? Due to space limit, we defer the comprehensive ablation study of \model{} to Appendix \ref{sec:ablation}.

\subsection{Reward and State Dynamics Clustered Environments}
\label{sec:reward_env}

\noindent\textbf{Environment setup.} We evaluated \model{} on two continuous control tasks with \emph{clustered reward functions}, simulated by MuJoCo \citep{todorov2012mujoco}. In \textbf{Ant-Goal}, the ant robot needs to move to a predetermined goal position. We created 4 clusters by distributing the goal positions in 4 different centered areas. In \textbf{Humanoid-Dir}, the human-like robot is controlled to move towards different target directions. We created 4 clusters by distributing target directions along 4 farthest apart directions in a 2D space. We also created environments with \emph{clustered transition functions} by adopting two movement environments \textbf{Hopper-Rand-Params} and \textbf{Walker-Rand-Params}, also simulated by MuJoCo. The physical parameters of the robot, including \emph{body mass}, \emph{damping on degrees of freedom}, \emph{body inertia} and \emph{geometry friction}, were manipulated to realize different transition functions of the robot's movement. The hopper and walker robots are required to move smoothly under different parameter settings. We created 4 clusters by manipulating one of the parameters at a time and keeping the others to the default parameters. The detailed procedure of task generation can be found in Appendix \ref{app:env}.

\begin{figure}[t]
    \begin{subfigure}{0.5\textwidth}
        \centering
        \includegraphics[height=3.5cm]{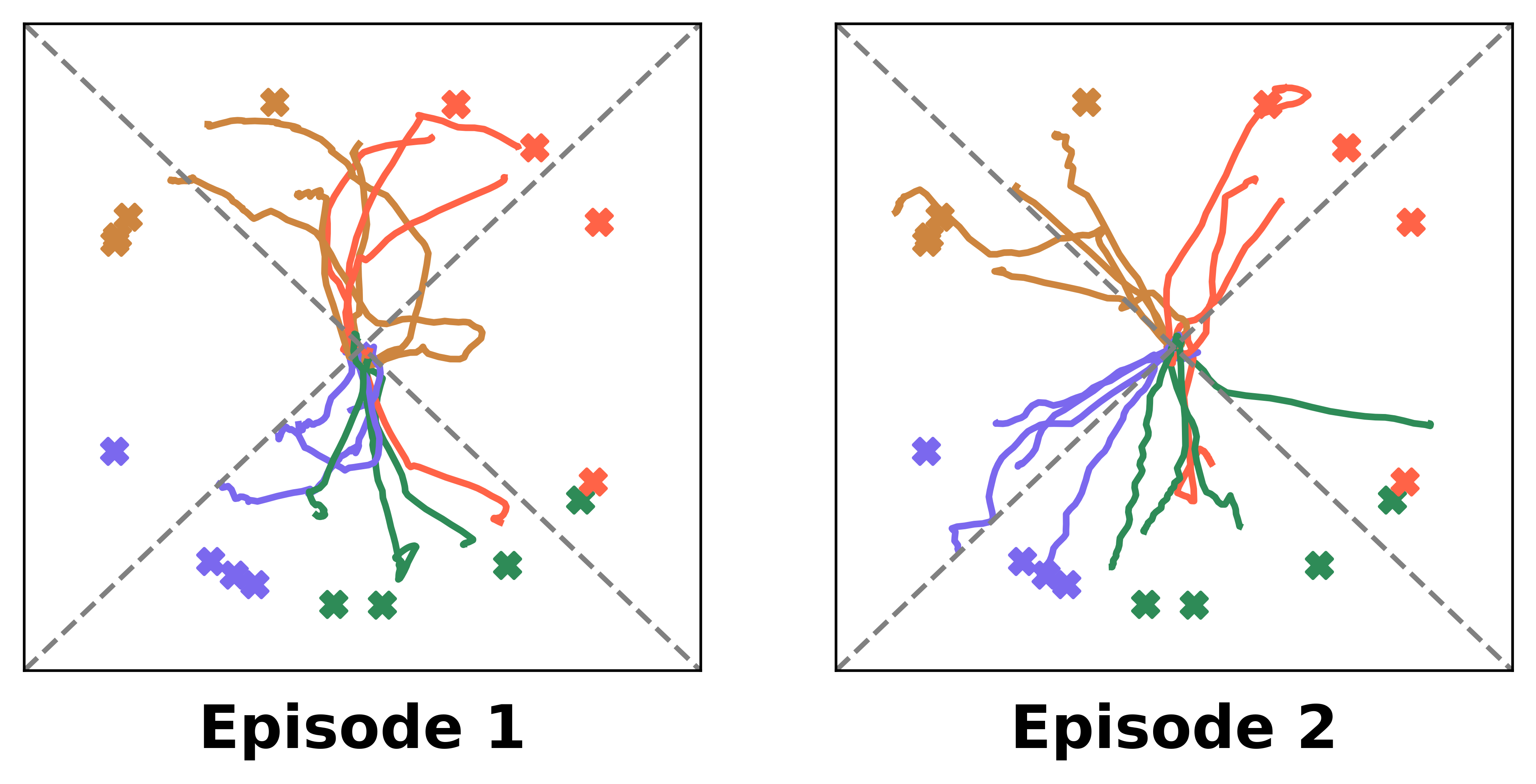}
        \caption{MiLEt traces.}
        \label{fig:milet_trace}
    \end{subfigure}
    \begin{subfigure}{0.5\textwidth}
    \centering
    \includegraphics[height=3.5cm]{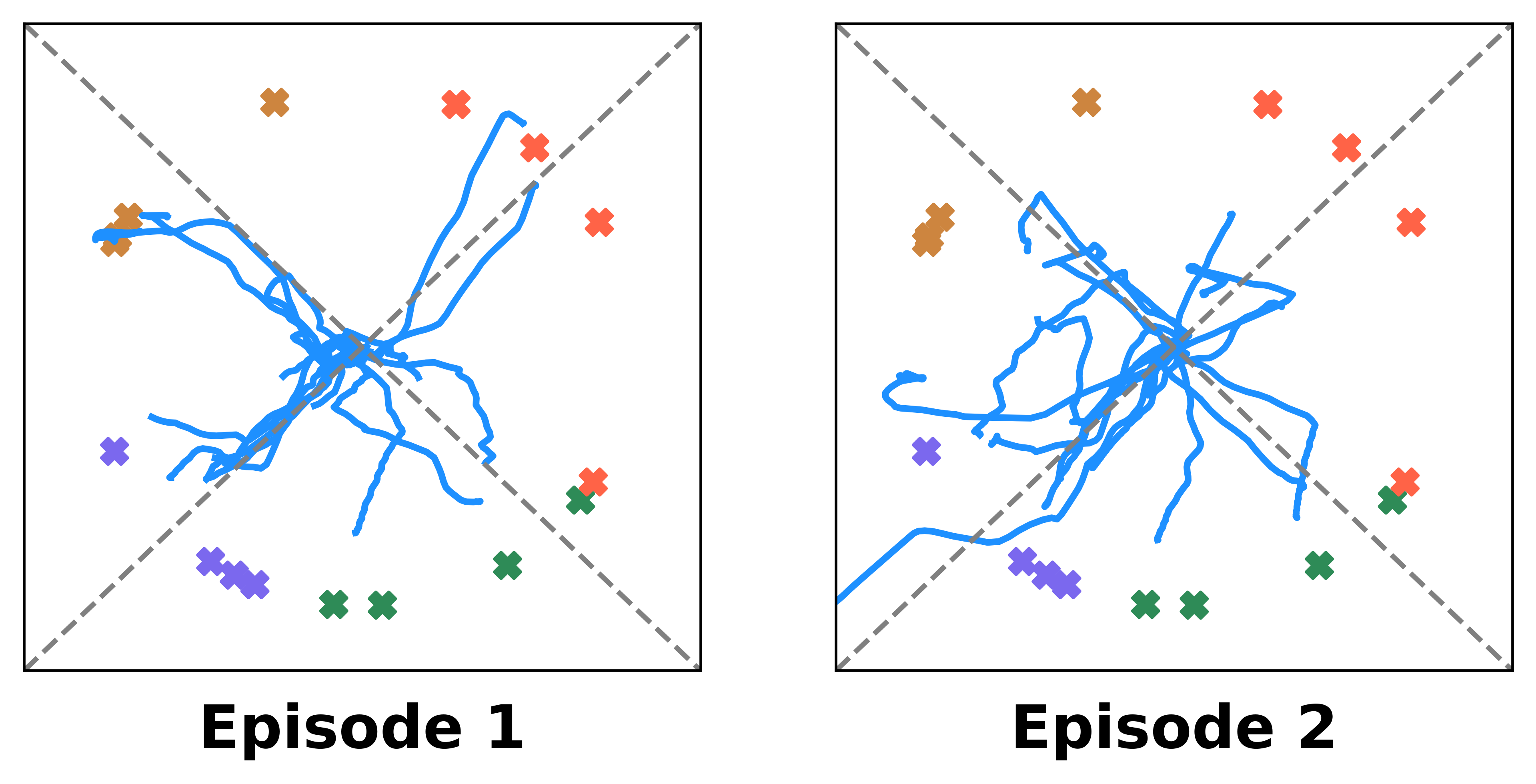}
    \caption{VariBAD traces.}
    \label{fig:varibad_trace}
    \end{subfigure}
    
    \begin{subfigure}{1\textwidth}
    \centering
    \includegraphics[height=4cm]{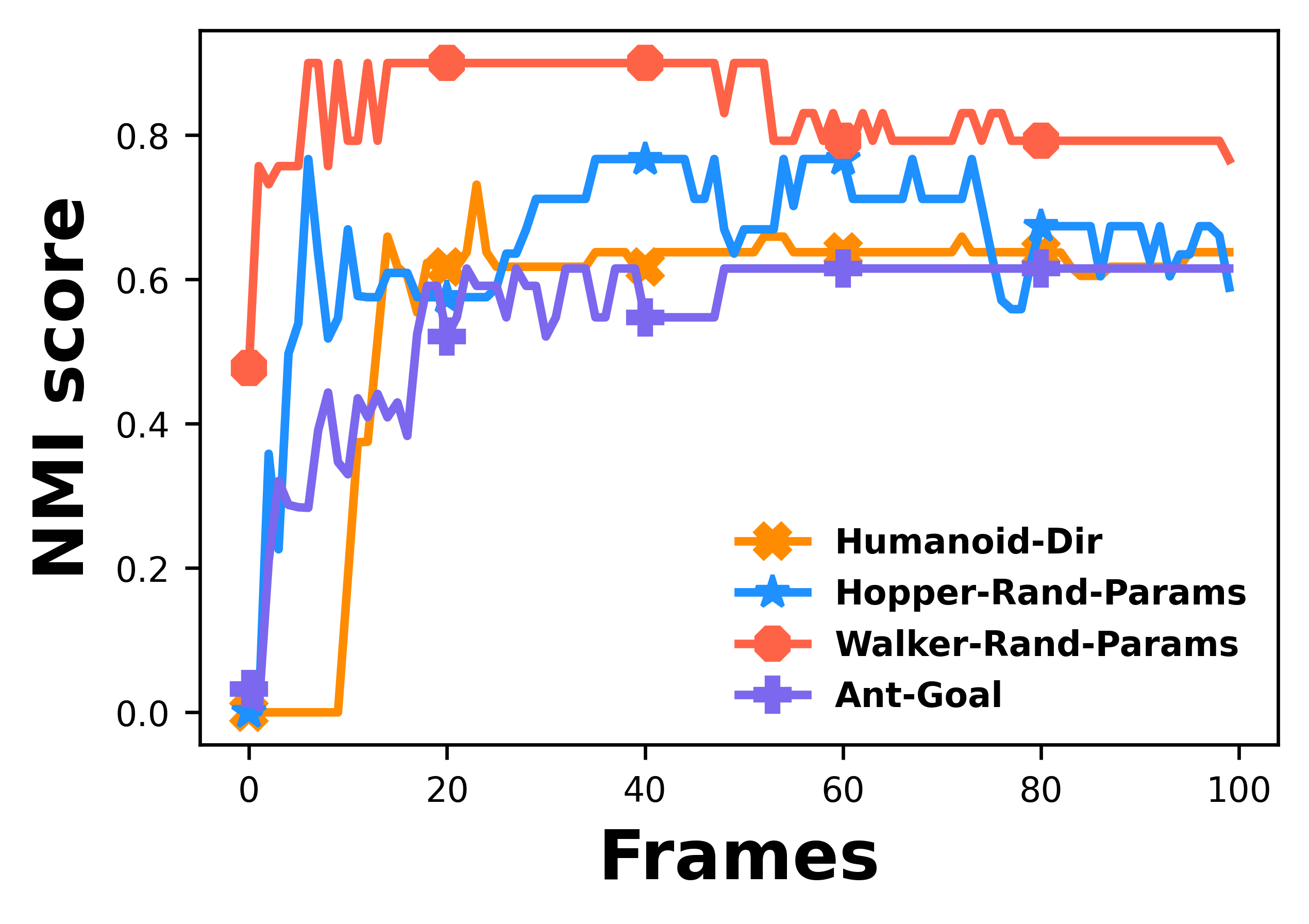}
    \caption{NMI score in exploration.}
    \label{fig:nmi}
    \end{subfigure}
    \vspace{-3mm}
    \caption{Qualitative analysis of \model{}. (a) Traces of \model{} on the meta-test tasks of Ant-Goal. Cross marks represent goal positions, and the colors represent the cluster assignments produced by \model{}. The dashed lines suggest the optimal traces to the centers of ground-truth clusters. (b) Traces of VariBAD on the same meta-test tasks of Ant-Goal. The traces are in the same color as VariBAD is unaware of clusters. (c) NMI of \model{}'s inferred clusters in the exploration episode of meta-test tasks in four environments.}
    \label{fig:behavior}
    \vspace{-2mm}
\end{figure}

\begin{figure*}[!tp]
        \centering
        \includegraphics[width=14.5cm]{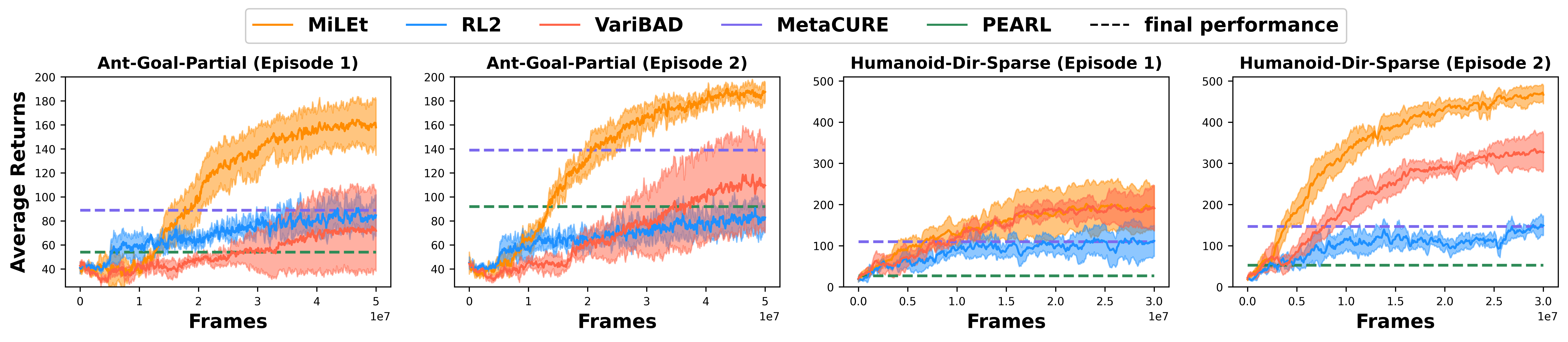}
        \vspace{-3mm}
        \caption{Average test performance for 2 episodes on sparse reward environments.}
        \label{fig:sparse}
        \vspace{-4mm}
\end{figure*}

\noindent\textbf{Baseline setup.} We compared \model{} with several representative meta-RL baselines, including RL$^2$ \citep{duan2016rl}, PEARL \citep{rakelly2019efficient}, VariBAD \citep{Zintgraf2020VariBAD:} and MetaCURE \citep{zhang2021metacure}. 
For each environment, we created 500 tasks for meta-train and hold out 32 new tasks for meta-test. We report the performance on test tasks during the meta-train phase. In the meta-test phase, we executed 2 episodes in each new task. For algorithms with an explicit exploration policy, i.e., \model{} and MetaCURE, we run their exploration policy in the first episode and exploitation policy in the second episode. We used public implementations of these algorithms provided by the original papers. We trained \model{} via Proximal Policy Optimization (PPO) \citep{schulman2017proximal} and set the default cluster number $C$ to 4. Because PEARL and MetaCURE are based on off-policy algorithms \citep{haarnoja2018soft}, they need less frames of data to converge in meta-train. We terminated them once the algorithm was converged and reported the final performance. We report the averaged performance over 3 random seeds. More hyper-parameter settings and implementation details can be found in Appendix \ref{app:hyper}.

\noindent\textbf{Results and analysis.} Figure \ref{fig:cluster_result} shows the test performance of all evaluated meta-RL algorithms. We also provide qualitative analysis in Figure \ref{fig:behavior}, including visualization of the models' behaviors and the clustering performance of \model{} in the exploration episode, measured by the normalized mutual information score (NMI). 

\model{} showed significant improvement against baselines in the second episode in testing. Interestingly, we can observe even though the first episode of \model{} was reserved for exploration, it still performed comparably to other methods in all four different environment setups. In the first episode, \model{} behaved exploratorily to find the most probable cluster of the current task, and thus its traces in Figure \ref{fig:milet_trace} look like spirals from the starting point. VariBAD is also designed to explore by uncertainty in task inference, but its traces are close to random walk at the early stage, which is less effective. In Figure \ref{fig:nmi}, we can observe the NMI scores of the \model{}'s inferred tasks have almost converged in 20 steps, which means the cluster inference became stable in an early stage and can thereby provide the agent helpful cluster-level information to gain fine-grained task information. 
This also explains how \model{} obtained comparable performance in the first episode. 
In the second episode, with cultivated task information, \model{} is able to move towards the targets directly, showing significant improvements against baselines. MetaCURE guides the exploration by task IDs, which in fact provides more information of environment than what \model{} can access. However, the exploration empowered by task IDs does not explicitly explore the coarser but useful information at the cluster level. \model{} is built based on the divide-and-conquer principle, which is more efficient to utilize such information by explicitly exploring clustering structures. 


\subsection{Sparse Reward Environments}
\noindent\textbf{Environment setup.} We evaluated \model{} on a more challenging setting where the reward is sparse. We modified Ant-Goal and Humanoid-Dir environments such that the agent only gets positive rewards within a small region around target positions or target directions, otherwise 0. We denote them as \textbf{Ant-Goal-Partial} and \textbf{Humanoid-Dir-Sparse}. 
In Ant-Goal-Partial, we found all evaluated methods failed when the rewards were too sparse. Hence, we set the reward regions to cover the initial position of the robot to warm start the agents. With sparse rewards, it becomes more important for the agent to leverage knowledge across related tasks, as the feedback within a single task is insufficient for policy learning. 
More environment details are deferred to Appendix \ref{app:env}.

\noindent\textbf{Results and analysis.} We present results in Figure \ref{fig:sparse}. By exploring and leveraging task structures in the exploration episode, \model{} outperformed all baselines with a large margin. MetaCURE is designed for exploration when reward is sparse by utilizing task IDs, which indeed helped it outperform other baselines in Ant-Goal-Partial. But such exploration is at the task level, and thus it is unable to effectively explore cluster information to enhance task modeling.
On the contrary, \model{} leveraged relatedness among tasks within the same cluster to bootstrap policy adaptation; and as previously shown in Figure \ref{fig:nmi}, cluster inference can be efficiently solved, which provides an edge for accurate task inference. These two factors contribute to \model{}'s better performance in the exploitation episode. 

\begin{table}[t]
\vspace{-2mm}
\centering
\caption{Performance comparison on Ant-Goal and Ant-U.}\label{exp:cluster}
\begin{tabular}{ccc}
\hline
 & Ant-Goal & Ant-U \\ \hline
VariBAD &    -168.6{\small $\pm 9.6$}        &  -162.4{\small $\pm 9.2$}      \\ \hline
\model{-2} &     -132.3{\small $\pm 7.6$}     &      -128.6{\small $\pm 8.8$}       \\ \hline
\model{-4} &   -125.4{\small $\pm 5.1$}    &    -113.7{\small $\pm 4.8$}      \\ \hline
\model{-6} &    -123.6{\small $\pm 4.4$}      &       -99.7{\small $\pm 5.2$}     \\ \hline
\model{-8} &   -124.2{\small $\pm 4.7$}     &      -117.9{\small $\pm 5.7$}     \\ \hline
\model{-10} &   -128.6{\small $\pm 5.2$}      &    -142.7{\small $\pm 10.4$}     \\ \hline
\end{tabular}
\vspace{-5mm}
\end{table}

\subsection{Influence from the Number of Clusters}
We also studied how the number of clusters $C$ set by the agent influences the final performance, especially when there is a mismatch between the ground-truth cluster size and $C$ set by the agent. We set $C$ to different values and denote it in suffixes of \model{}. We additionally created a set of tasks on the Ant-Goal environment, where the goal positions were uniformly sampled on a circle. We denote it as \textbf{Ant-U}. In this setting, there are no explicitly clusters in prior task distribution. 

\begin{figure}[t]
    \begin{subfigure}{0.5\textwidth}
        \centering
        \includegraphics[height=3.5cm]{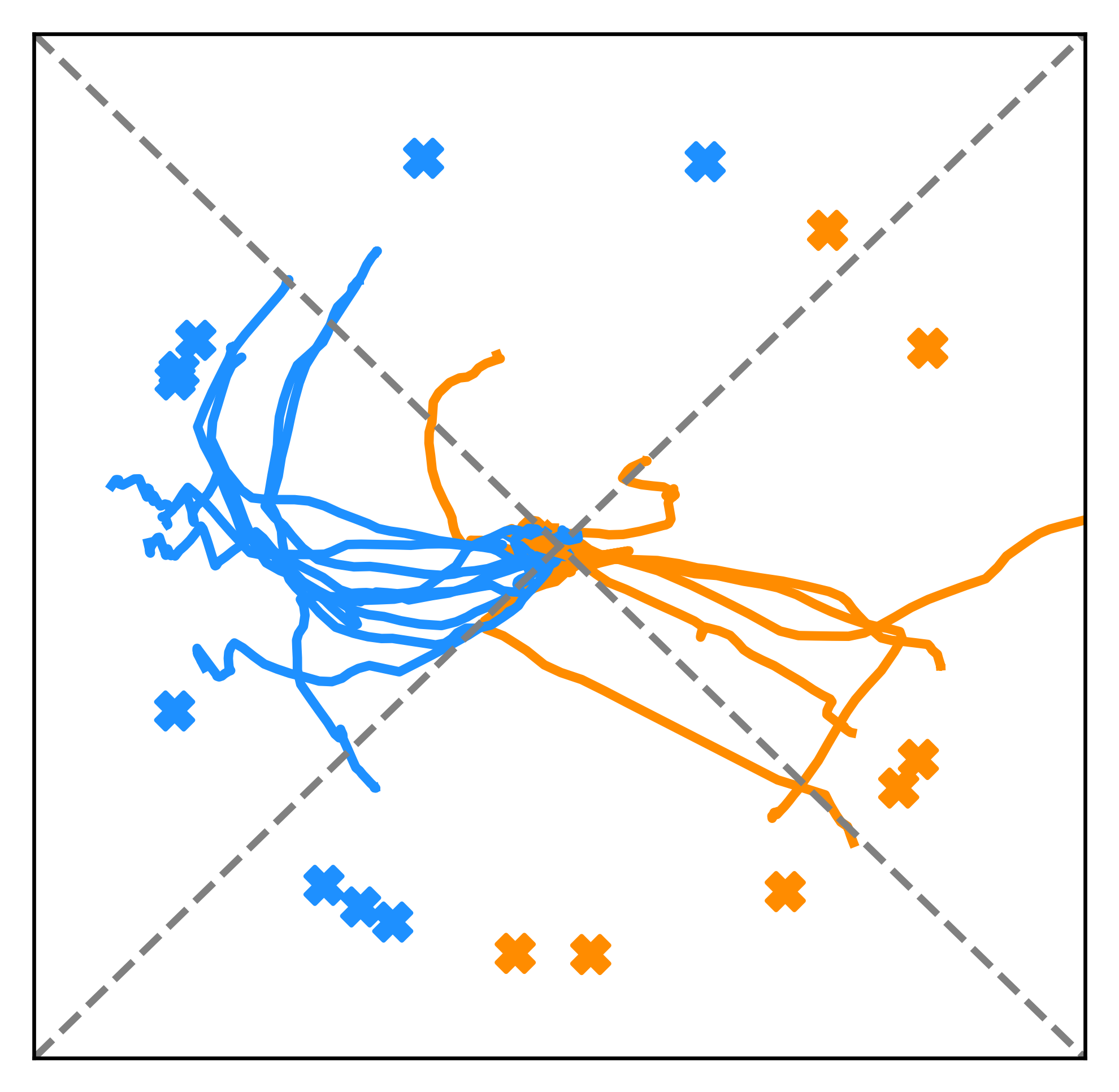}
        \caption{\model{-2} on Ant-Goal.}
        \label{fig:milet-2}
    \end{subfigure}
    \begin{subfigure}{0.5\textwidth}
        \centering
        \includegraphics[height=3.5cm]{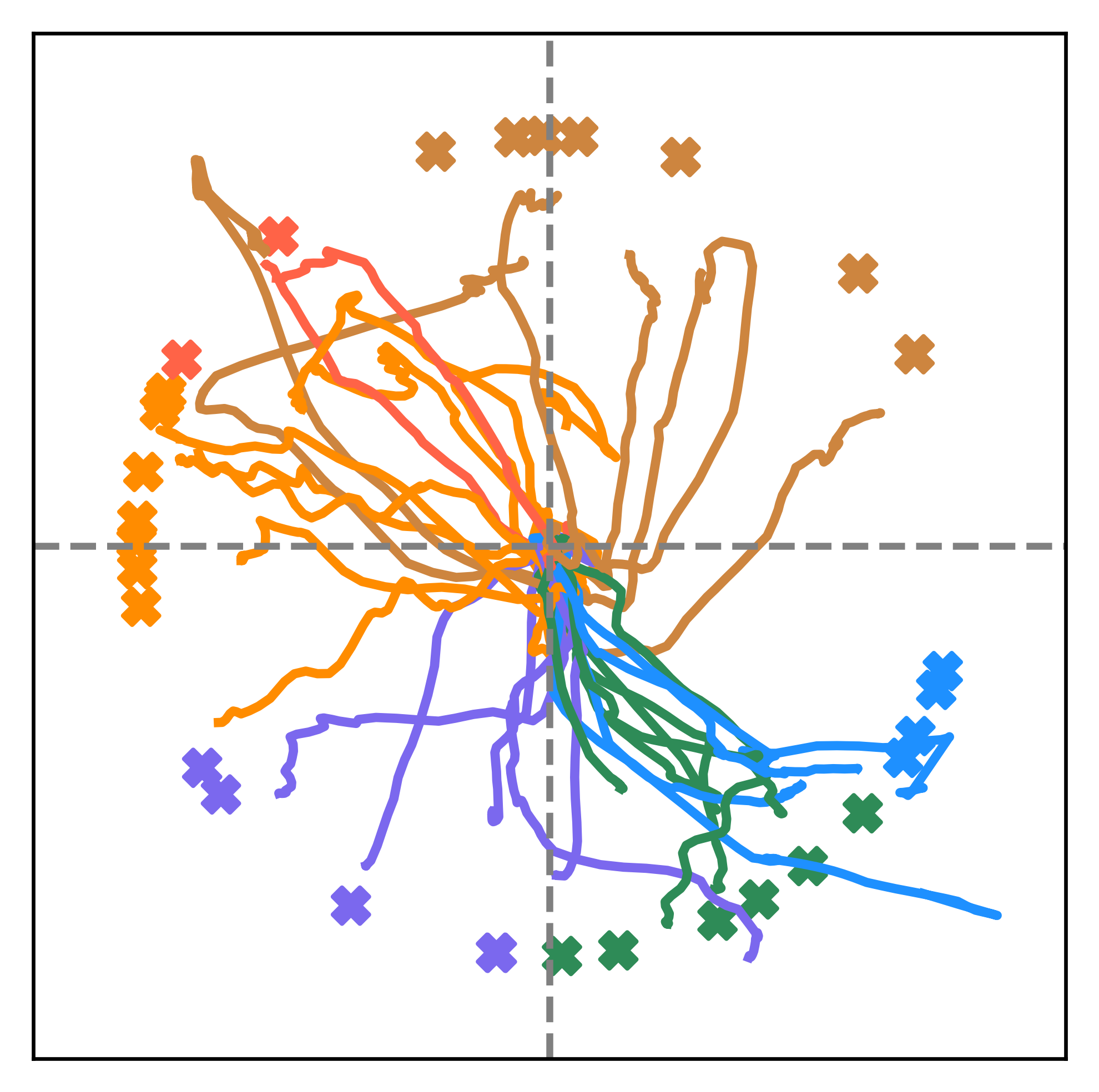}
        \caption{\model{-6} on Ant-U.}
        \label{fig:milet-6}
    \end{subfigure}
    \vspace{-2mm}
    \caption{Traces of \model{-2} and -6 in exploration episodes. Colors represent the cluster assignments produced by \model{}.}
    \label{fig:milet-26}
    \vspace{-2mm}
\end{figure}

The average final returns are shown in Table \ref{exp:cluster}. Interestingly, we observe \model{} can perform well even though there is no explicit cluster structure in Ant-U. By looking into the detailed trajectories, we found \model{} segmented the circle into different parts as shown in Figure \ref{fig:milet-6} such that knowledge from nearby tasks can be effectively shared. VariBAD mistakenly assumed all tasks can share knowledge and thus failed seriously. 
 When $C$ is set smaller than the ground-truth number of clusters, \model{-2} discovered more general structures (as shown in Figure \ref{fig:milet-2}). However, transferable knowledge within such structures is limited as distinct clusters are merged, causing the performance drop. However, it does not mean more clusters than necessary is helpful, as less knowledge could be shared in each cluster. In Ant-U, \model{-8} and -10 generated unnecessary clusters, and cluster assignments are mixed at the boundary of adjacent clusters (see visualization in Appendix \ref{app:vis_ant_u}). Such inaccurate cluster modeling causes ineffective exploration and knowledge sharing, leading to degenerated performance.

\subsection{Results on Meta-World Environments}
\begin{figure}[t]
    \vspace{-1mm}
    \centering
    \includegraphics[width=10cm]{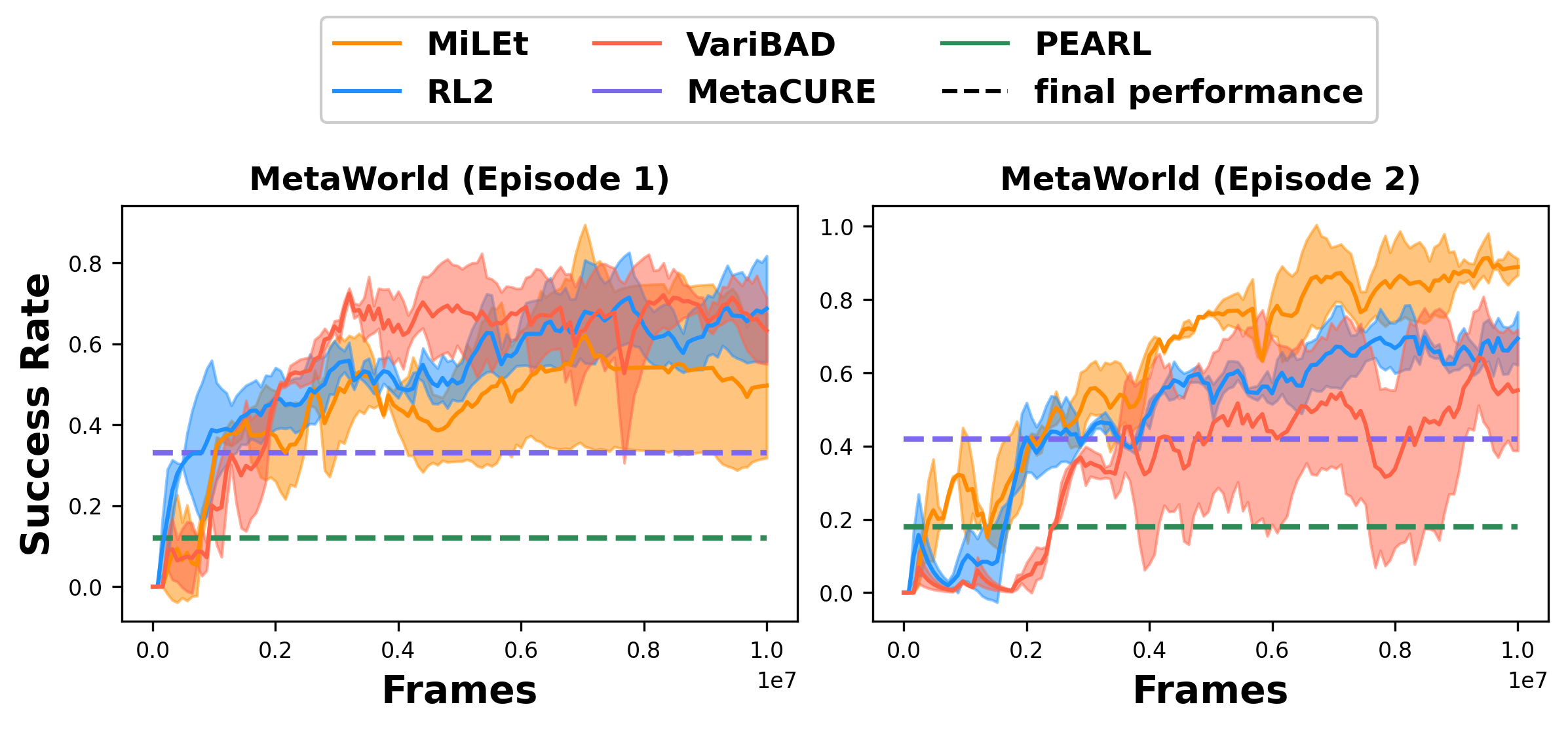}
    \caption{Average test performance for 2 episodes on Meta-World environments.}
    \label{fig:metaworld}
    \vspace{-4mm}
\end{figure}

We also evaluated \model{} on a challenging meta-RL benchmark Meta-World \citep{yu2020meta}, including a variety of robot arm control tasks. Each task needs specific skills to solve, e.g., pushing and pulling, and different tasks share different degrees of relatedness. 
There are also variants inside each task, e.g., positions of objects and targets. We considered a combination of two set of tasks: Plate-Slide and Plate-Slide-Side, where the initial position of the arms and objects are the same and hence disclose less information about tasks to the agents upfront. The former needs the agent to hold the plate and move it to the cabinet and the latter needs the agent to push the plate into the cabinet sideways. We held out 8 variants of each task for testing
and present results in Figure \ref{fig:metaworld}. \model{} achieved higher success rates in the second episode based on its obtained task information from the first episode.

\section{Conclusion \& Future Work}
In this paper, we present \model{}, a clustering-based solution to discover structured heterogeneity of tasks in meta-RL. \model{} is able to discover clustered task structures in a population of RL-tasks and adapt cluster-level transferable information to new tasks. To quickly identify the cluster assignment of new tasks, \model{} learns a separate exploration policy which aims to rapidly reduce uncertainty in cluster inference when interacting with the environment. We further design a dedicated reward function to control the exploration between the cluster- and task-level information. 
\model{} outperformed representative meta-RL baselines in a variety of empirical evaluations. 

\model{} sheds light on discovering structures in the task distribution to boost meta-RL. Several extensions of \model{} are worth exploring in the future work. We currently assume a uniform cluster prior. More complicated priors can be considered to enable new features. For example, the Dirichlet process prior can be used to automatically identify number of clusters in the task distribution, and possibly detect out-of-distribution tasks in the meta-test phase. Also, \model{} can be combined with skill-based RL methods \citep{pertsch2021accelerating, nam2022skillbased}  to learn cluster-level skills, which will form a new basis for meta-RL, e.g., each task can be modeled as a mixture of skills and different clusters of tasks associate with different skill distributions. 

\bibliography{sample}

\newpage
\appendix
\onecolumn
\newpage
\section{ELBO derivation}
\label{app:elbo}
\begin{align*}
   \mathbb{E}_{\rho(\mathcal{M}, \tau^+)}\Big[\log p(\tau^+)\Big] &= \mathbb{E}_{\rho} \Big[ \ln \int p(z, c, \tau^+) \frac{q(z,c|\tau_{:t})}{q(z,c|\tau_{:t})} \Big] \\
   &= \mathbb{E}_{\rho} \Big[  \ln \mathbb{E}_{q(z,c|\tau_{:t})}\Big[ \frac{p(z, c, \tau^+)}{q(z,c|\tau_{:t})} \Big]\Big] \\
   &\geq \mathbb{E}_{\rho}\Big[\mathbb{E}_{q(z,c|\tau_{:t})}\Big[ \ln \frac{p(z, c, \tau^+)}{q(z,c|\tau_{:t})}  \Big] \Big] \\
   &= ELBO_t
\end{align*}
where the inequality holds due to the Jensen's inequality. The inner expectation term can be further decomposed as, 
\begin{align*}
&~\mathbb{E}_{q(z, c|\tau_{:t})}\Big[ \ln\frac{p(z, c, \tau^+)}{q(z, c|\tau_{:t})} \Big] \\
   = & ~\mathbb{E}_{q(z, c|\tau_{:t})}\Big [\ln \frac{p(\tau^+|z, c)p(z|c)p(c)}{q(z|c, \tau_{:t})q(c|\tau_{:t})} \Big ] \\
   = & ~\mathbb{E}_{q(z, c|\tau_{:t})} \Big[\ln p(\tau^+|z, c) + \ln \frac{p(z|c)}{q(z|c, \tau_{:t})} + \ln \frac{p(c)}{q(c|\tau_{:t})}  \Big ] \\
    =&  ~\mathbb{E}_{q(z, c|\tau_{:t})} \Big[\ln p(\tau^+|z, c) \Big ] - \mathbb{E}_{q(c|\tau_{:t})} \Big[ \text{KL}(q(z|c, \tau_{:t}) \parallel p(z|c)) \Big]
    - \text{KL}(q(c|\tau_{:t})\parallel p(c)), 
\end{align*}
 which is essentially Eq.(\ref{eq:elbo}). In practice, we first apply the Gumbel-Softmax trick \citep{jang2017categorical} to sample a $\Tilde{c}$ to calculate the second term instead of directly calculating the expectation, and then sample a $\Tilde{z}$ given $\Tilde{c}$ to calculate the first term using the reparameterization trick \citep{kingma2013auto}. To calculate the last term, we decompose it into,
\begin{align*}
     -\text{KL}(q(c|\tau_{:t})\parallel p(c)) = H\big[ q(c|\tau_{:t})\big] +\mathbb{E}_{q(c|\tau_{:t})}\big [p(c)\big ],
\end{align*}
where $p(c)$ is the uniform prior of the cluster distribution, thus the last KL term is essentially the entropy of the posterior $q(c|\tau_{:t})$ plus a constant value, which can be easily computed.

\section{Algorithms}
\label{sec:alg}

\begin{algorithm}[hpt]
\SetAlgoLined
\Require A set of meta-train tasks $\mathcal{M}$ drawn from $p(\mathcal{M})$\; 

Initialize a buffer $\mathcal{V}$ for CBVI training\;

 \While{not \textup{Done}}{
    Sample a task $M_i\sim \mathcal{M}$\;
    
    Collect exploration and exploitation episodes $\tau^+=\{\tau^\psi, \tau^\phi\}$ by running Alg.(\ref{alg:meta_test}) on $M_i$\;
   
    Insert $\tau^+$ to $\mathcal{V}$\;
    
    Compute intrinsic rewards $r_h$ and $r_c$\;
    
    Train $\pi_\psi$ on $\tau^\psi$ by maximizing Eq.(\ref{eq:exploration_reward})  using PPO\;
    
    Train $\pi_\phi$ on $\tau^\phi$ by maximizing Eq.(\ref{eq:expected_reward}) using PPO\;
    
    Sample a trajectory batch $v$ from $\mathcal{V}$\;
    
    Update $\theta, \omega$ using $v$ by maximizing Eq.(\ref{eq:vae_loss})\;
 }
 \caption{\model{}: Meta-train Phase}
 \label{alg:meta_train}
\end{algorithm}

\begin{algorithm}[hpt]
\SetAlgoLined
\Require Meta-test task drawn from $p(\mathcal{M})$, number of exploitation episodes $N$\;

\For{$t = 1, ..., H$}{
    Obtain $c = \argmax_c q_\theta(c|h_{\alpha, t-1}^\psi)$ \tcp*{Use Gumbel-Softmax during training}

    Take action according to $\pi_\psi\big (a_t|s_t, q_\theta(z|h_{\beta, t-1}^\psi, c) \big )$  \;
    
    Update $h_t^\psi$ with $(s_t, a_t, r_t, s_{t+1})$\; 
}

Initialize $h_0^\phi = h_H^\psi$ \;

\For{$e=1,..., N$}{
    \For{$t = 1, ..., H$}{
    Obtain $c = \argmax_c q_\theta(c|h_{\alpha, t-1}^\phi)$  \tcp*{Use Gumbel-Softmax during training}

    Take action according to $\pi_\phi\big (a_t|s_t, q_\theta(z|h_{\beta, t-1}^\phi, c) \big )$\; 
    
    Update $h_t^\phi$ with $(s_t, a_t, r_t, s_{t+1})$\; 
    }
}
 \caption{\model{}: Meta-test Phase}
 \label{alg:meta_test}
\end{algorithm}

Algorithm \ref{alg:meta_train} summarizes the training procedure of \model{}. We use the on-policy algorithm PPO \citep{schulman2017proximal} to train policies; and retain a buffer storing trajectories for the cluster-based variational inference training. Algorithm \ref{alg:meta_test} is used to obtain trajectories given a task. When it is executed in Algorithm \ref{alg:meta_train} during training, $c$ is sampled using the Gumbel-Softmax trick, otherwise using argmax when handing new tasks during testing.

\section{Environments}
\label{app:env}
\noindent\textbf{Ant-Goal.} We create clusters by manipulating the goal positions. Given an angle $\theta \in [0, 2]$, we obtain the 2D coordinate of the goal position as $(r\cos \theta, r\sin \theta)$, where $r$ is the radius and fixed to 2 in our experiments. To create clusters, we sample $\theta$ from 4 different normal distributions $\mathcal{N}(0.25, 0.2^2)$, $\mathcal{N}(0.75, 0.2^2)$, $\mathcal{N}(1.25, 0.2^2)$, $\mathcal{N}(1.75, 0.2^2)$.  To sample a task, we first uniformly sample a normal distribution, and then sample a $\theta$ from it.

\noindent\textbf{Humanoid-Dir.} We create clusters by manipulating the goal directions. To sample a goal direction, we first sample an angle $\theta \in [0, 2]$, and then the goal direction is $(\cos \theta, \sin \theta)$. To create clusters, we sample $\theta$ from 4 different normal distributions $\mathcal{N}(0.25, 0.2^2)$, $\mathcal{N}(0.75, 0.2^2)$, $\mathcal{N}(1.25, 0.2^2)$, $\mathcal{N}(1.75, 0.2^2)$. To sample a task, we first uniformly sample a normal distribution, and then sample a $\theta$ from it.

\noindent\textbf{Hopper-Rand-Params.} We create clusters by manipulating physical parameters of the robot, causing different transition functions. We have four sets of parameters in total, including \emph{body mass}, \emph{damping on degrees of freedom}, \emph{body inertia} and \emph{geometry friction}. To sample a task, we first uniformly sample one of the four parameter sets, and then multiply parameters in it with multipliers sampled from $\mathcal{N}(3, 1.5^2)$.

\noindent\textbf{Walker-Rand-Params.} The parameter sets are the same as the hopper robot. We find small multipliers cannot change transition functions a lot in this environment, thus we sample the multipliers from $\mathcal{N}(6, 1^2)$.

\noindent\textbf{Ant-Goal-Partial.} The tasks are sampled in the same way as Ant-Goal. The goal reward is defined as,
\begin{equation*}
    r=\left\{\begin{matrix}
-|x-g|_1 + t, & |x-g|_1 \leq t \\
0, & otherwise \\
\end{matrix}\right.
\end{equation*}
where $x$ is the current coordinate of the robot, $g$ is the coordinate of the goal position, $t$ is the threshold, whish is fixed to 3 in our setting. Control cost and contact cost are also included in the final reward. 

\noindent\textbf{Humanoid-Dir-Sparse.} The tasks are sampled in the same way as Humanoid-Dir. The velocity reward is defined as,
\begin{equation*}
    r=\left\{\begin{matrix}
g\cdot v, & \frac{g\cdot v}{\left \| g \right \|\left \| v \right \|} \geq t \\
0, & otherwise \\
\end{matrix}\right.
\end{equation*}
where $v$ is the velocity of the robot, $g$ is the goal direction, $t$ is the threshold and fixed to 0.8 in our setting. Control cost and contact cost are also included in the final reward.

\noindent\textbf{Ant-Goal-U.} For each task, we uniformly sample a $\theta \sim \mathcal{U}(0, 2)$ and obtain the goal position $(r\cos \theta, r\sin \theta)$ with $r$ fixed to 2.

\section{Hyper-parameters}
\label{app:hyper}
\begin{figure*}[!htp]
    \centering
    \includegraphics[width=7cm]{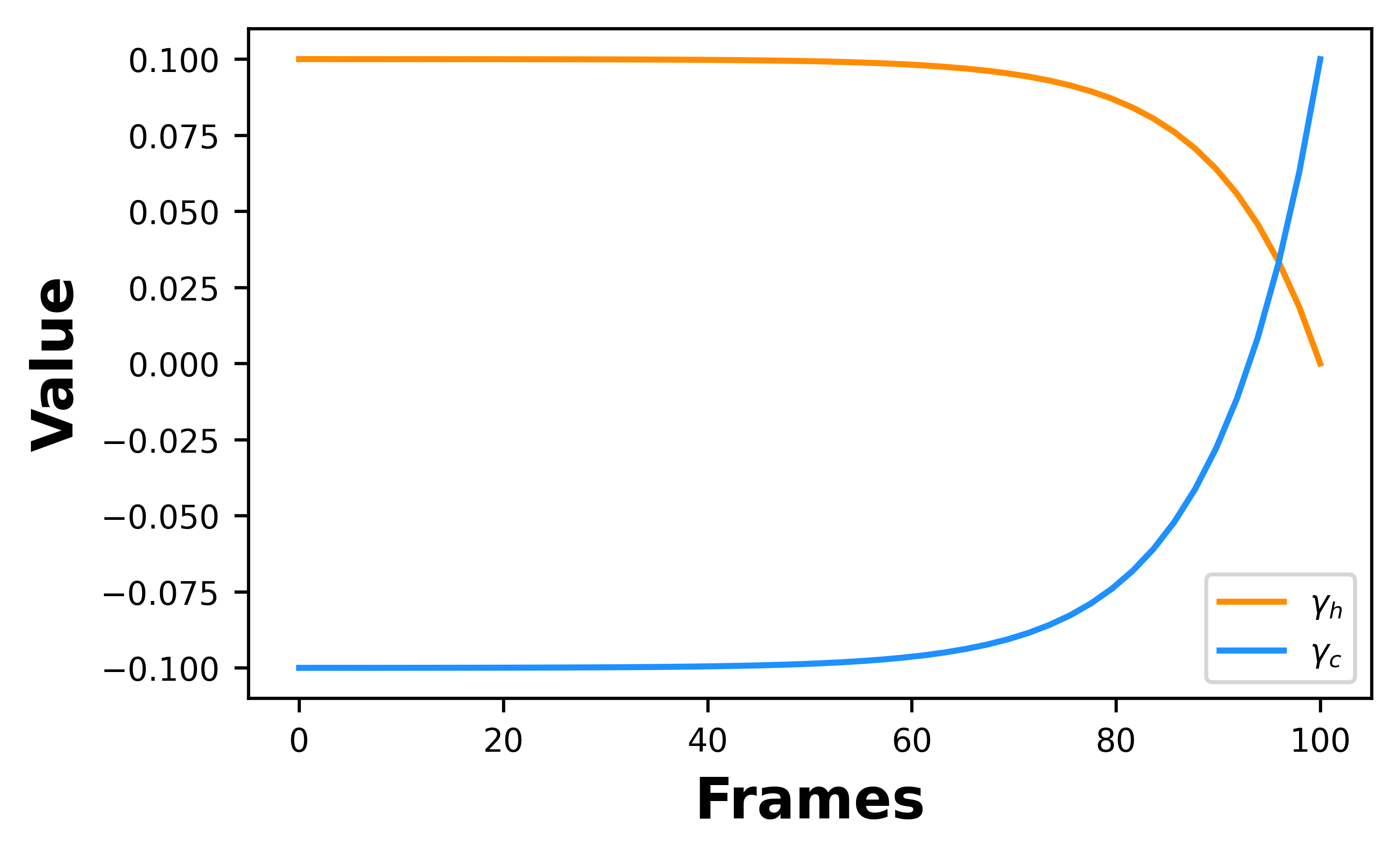}
    \caption{Visualization of $\gamma$ in the exploration policy.}
    \label{fig:gamma}
\end{figure*}

We used the PyTorch framework \citet{paszke2019pytorch} to implement our experiments. We use PPO with Humber loss to update \model{}. In the cluster-based variation inference, we set $\lambda_\text{I}$ to 1, and $\lambda_\text{P}$ to 0.1. We update the target network $p_\textup{tgt}(z|c)$ every 50 training epochs. For the decaying function used in exploration policy, we set $\gamma_h(t)=0.1-0.1\exp(-0.1(H-t))$ and $\gamma_c(t)=-0.1+0.2\exp(-0.1(H-t))$, which are visualized in Figure \ref{fig:gamma}. We use the Adam optimizer \citep{kingma2014adam} for both policy learning and cluster-based variational inference training. For policy learning including $\pi_\psi$ and $\pi_\phi$, we set the learning rate as 1e-4, while 1e-3 for variational inference training. In the clustered reward function environments, $\lambda_s$ is set to 0 in MiLEt and all variational inference baselines. In the clustered state transition function environments, $\lambda_s$ is set to be 1. The other parameters are set as default. The max episode length $H$ is set to 100 for every environment.

We provide implementation of \model{} and task generation process in the supplementary material. All experiments are run on an NVIDIA Geforce RTX 3080Ti GPU with 12 GB memory.

\section{Ablation Study}
\label{sec:ablation}
We study the contribution of proposed components in \model{}. Firstly, we disabled the exploration policy and directly train the exploitation policy to maximize task rewards. Secondly, we removed the stacked GRU (S-GRU) structure and only keep a single GRU to encode the trajectory $\tau_{:t}$ in the cluster-based variational inference. Thirdly, we disable the consistency regularizers (CR) in the variational inference and directly optimize the ELBO. Lastly, we set $\gamma_c(t) = 0$ to study the influence of the consistency reward for the exploration policy. We also include a variant of VariBAD by ablating the exploration policy and S-GRU of \model{}, named VariBAD-G. We use the same environments as Section \ref{sec:reward_env}.

The final performance of all variants is shown in Table \ref{tab:ablation}. Firstly,        VariBAD-G has the same neural network architecture of the actor-critic module and encoder-decoder network as in VariBAD, but the VAE part is replaced by the Gaussian mixture VAE to model the clustering structures of tasks. It outperformed the baselines in general, showing that modeling the cluster structures is important to utilize structured heterogeneity in the task distribution, and our performance gain is not simply from the superior neural network architectures. Secondly, without the exploration policy, the performance decreases significantly. This variant can still unravel clustering structures of tasks. However, it cannot efficiently identify the clustering structures as the exploration does not aim to explore different cluster structures (also supported by low NMI score in Table \ref{tab:nmi_explore}). Our exploration policy is trained to explore the most certain clusters for tasks, leading to better adaptation performance. Secondly, removing S-GRU makes it harder to utilize patterns with different levels of granularity in the interaction history. This ability is essential to quickly identify cluster structures as cluster patterns are more common across tasks, without which the exploration is less effective, showing a lower NMI score in Table \ref{tab:nmi_explore}. Also, a consistent variational inference is very important. Our designed consistency regularizers are helpful to discover cluster structures following the nature of sequential data in meta-RL, improving the final performance of \model{}. Finally, after setting $\gamma_c(t) = 0$, the performance drops in general. A negative $\gamma_c(t)$ encourages the policy to explore different clusters, without which the exploration could stick on imperfect cluster assignments encouraged by $r_h$.

\begin{table}[!htp]
\caption{Ablation analysis of \model{}.}
\label{tab:ablation}
\centering
\begin{tabular}{c|c|c|c|c}
\hline
 & Ant-Goal & Humanoid-Dir & Hopper-Rand-Params & Walker-Rand-Params\\ \hline
 VariBAD-G  & -171.2{\small $\pm 9.7$}  &  514.3{\small $\pm 24.6$}  &  272.9{\small $\pm 11.3$}   &  153.6{\small $\pm 9.4$}  \\ \hline
$\neg$exploration & -130.4{\small $\pm 4.5$}    &   533.9{\small $\pm 18.4$}       &          278.3{\small $\pm 9.6$}          &       164.2{\small $\pm 16.7$}             \\ \hline
$\neg$S-GRU &    -173.7{\small $\pm 14.2$}     &       554.2{\small $\pm 12.7$}        &         284.4{\small $\pm 14.3$}           &     211.7{\small $\pm 9.8$}                \\ \hline
 $\neg$CR &       -142.8{\small $\pm 8.7$}       &       528.1{\small $\pm 35.6$}          &    301.6{\small $\pm 12.2$}   &     219.4{\small $\pm 11.1$}    \\ \hline
 $\neg r_c$   &  -134.5{\small $\pm 9.4$}    &    542.1{\small $\pm 11.8$}   &   287.2{\small $\pm 10.6$}   & 189.1{\small $\pm 7.8$}     \\ \hline
\model{} & -125.4{\small $\pm 5.1$}    &  577.7{\small $\pm 28.0$}       &            312.2{\small $\pm 18.1$}        &     232.6{\small $\pm 15.4$}                \\ \hline
\end{tabular}
\end{table}

\section{Visualization of State Transition Functions}
\begin{figure}[!htp]
    \begin{subfigure}{0.5\textwidth}
        \centering
        \includegraphics[height=4.4cm]{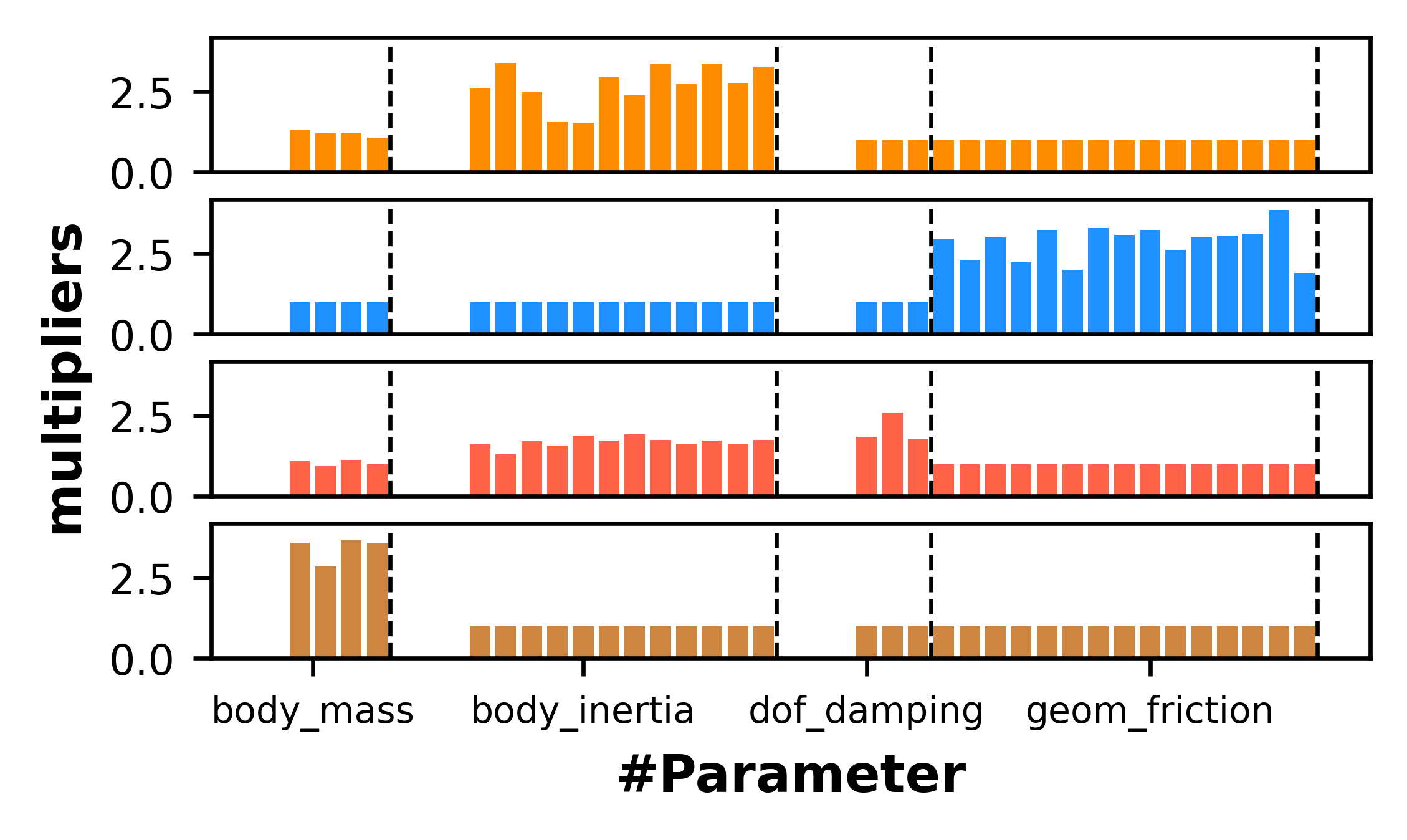}
        \caption{Hopper-Rand-Params.}
    \end{subfigure}
    \begin{subfigure}{0.5\textwidth}
    \centering
    \includegraphics[height=4.4cm]{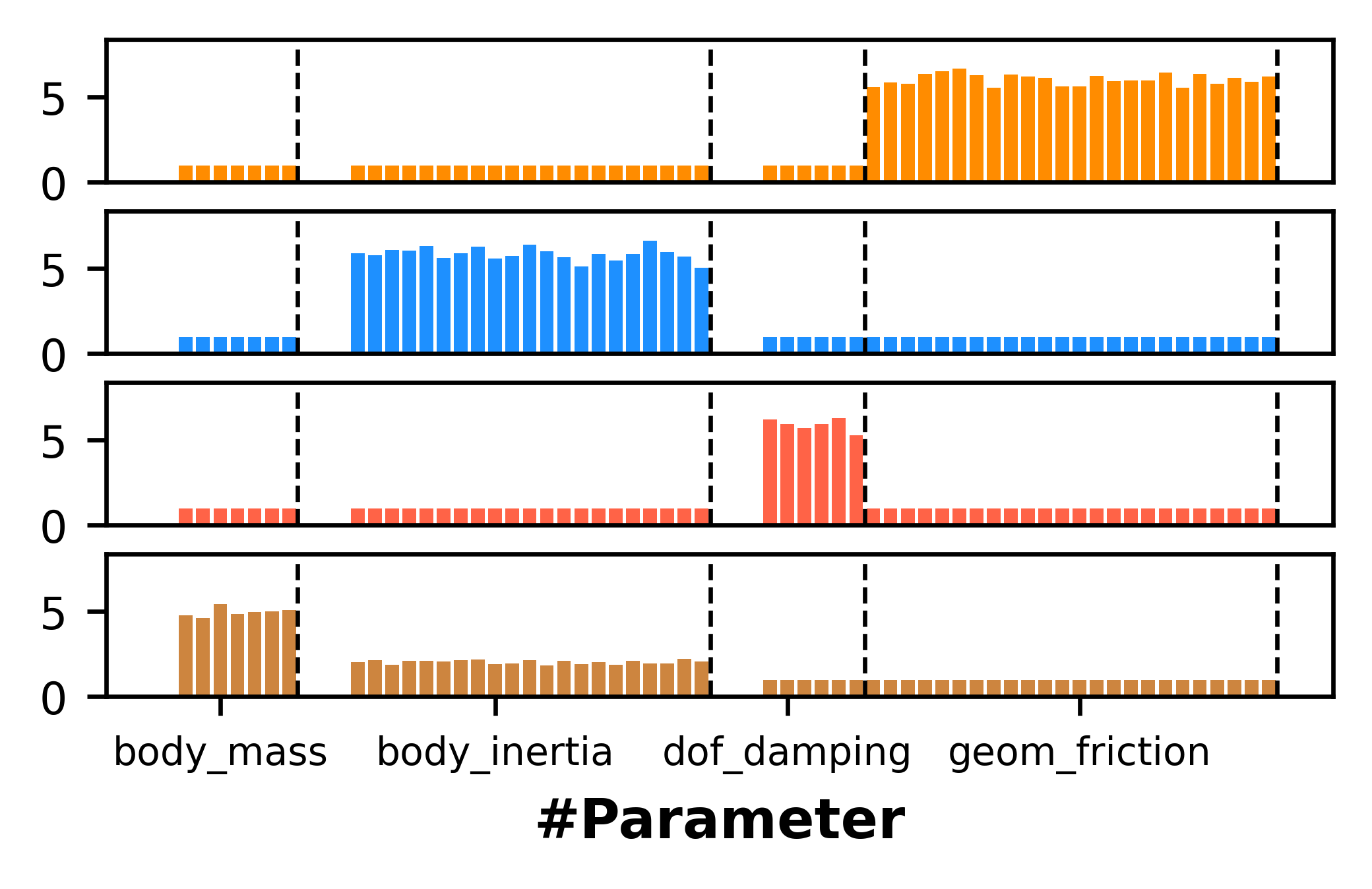}
    \caption{Walker-Rand-Params.}
    \end{subfigure}
    \caption{Clusters in state transition functions. Each color represents an identified cluster by \model{}. Four parameter sets are split by dashed lines. }
    \label{fig:cluster_trans}
\end{figure}

We visualize the inferred clusters in the clustered transition function experiments in Figure \ref{fig:cluster_trans}. It is hard to directly visualize the state transition functions, as they are not simple tabular functions. Recall that we created the ground-truth clusters by applying different multipliers on the initial parameters of each of the parameter sets in the corresponding environment, and thus such multipliers well characterize the state transition in each task. We then calculate the average multipliers of tasks in each inferred cluster \footnote{Multipliers with value 0 mean the corresponding initial parameters are 0.}. Ideally, one cluster of tasks should only have one set of multipliers larger than 1, as it is how we created different task clusters. From the visualization results, we can clearly observe that \model{} successfully identified the difference among the 4 ground-truth clusters.

\section{Visualization on Ant-U}
\label{app:vis_ant_u}
\begin{figure}[htp]
    \centering
    \includegraphics[width=8cm]{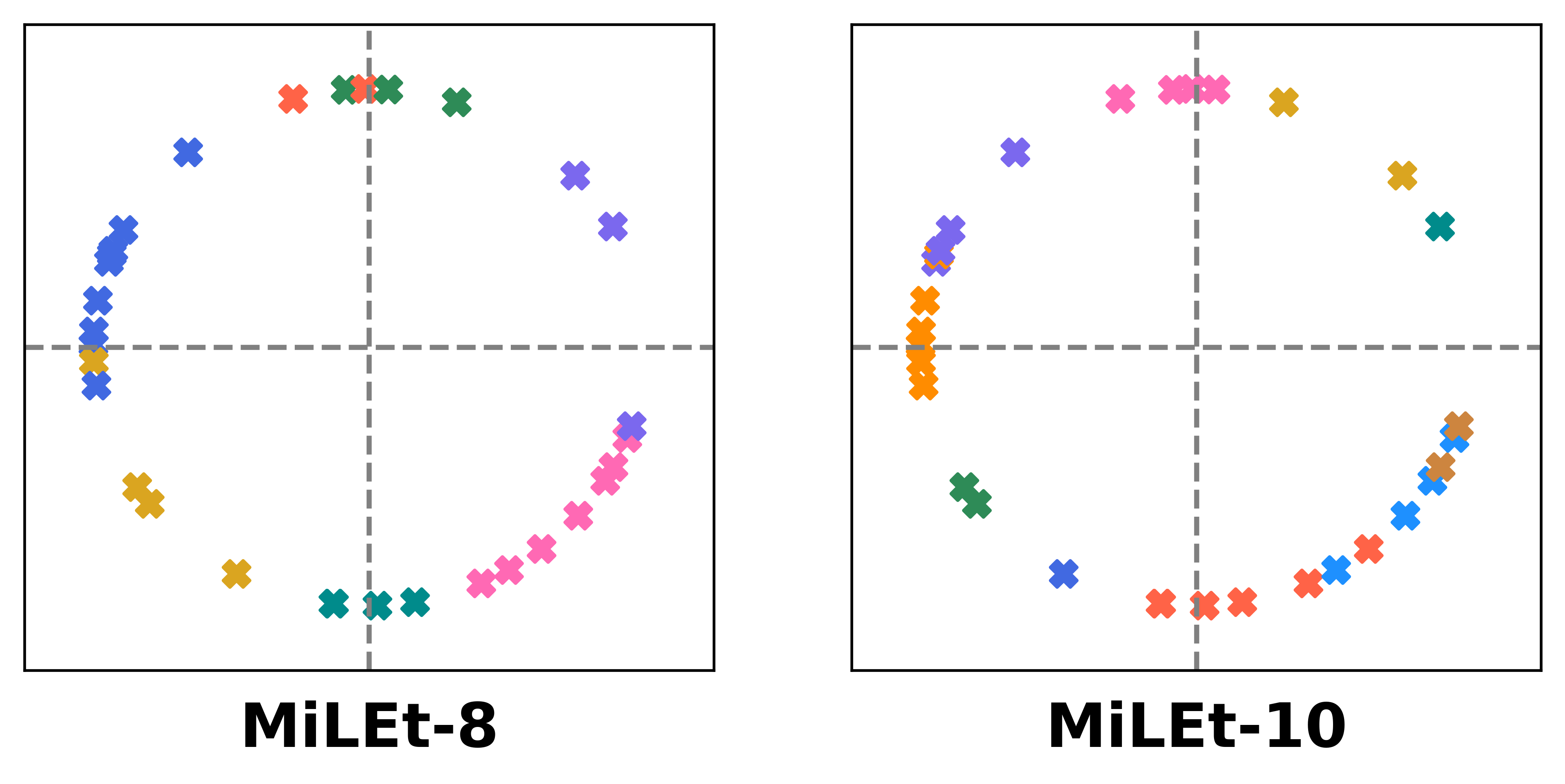}
    \caption{Cluster assignments produced by \model{-8} and \model{-10} in the Ant-U environment.}
    \label{fig:vis_ant_u}
\end{figure}

In Figure \ref{fig:vis_ant_u}, we visualize the task clusters identified by \model{-8} and \model{-10} on the testing tasks in the Ant-U environment. We can observe that tasks are split into smaller groups by these two new variants, and their cluster assignments are mixed at the boundary of adjacent clusters. Such inaccurate cluster modeling causes ineffective exploration and sharing of wrong knowledge across tasks, leading to the degeneration of final performance in Table \ref{exp:cluster}.

\section{Discussion of Cluster-aware Exploration}
Our exploration policy distinguishes \model{} from other task inference methods in utilizing the structures of the task distribution. VariBAD \citep{Zintgraf2020VariBAD:} performs exploration as a function of task uncertainty by maximizing task rewards. MetaCURE \citep{zhang2021metacure} and DREAM \citep{liu2021decoupling} propose to explore by maximizing the mutual information between inferred task embeddings and pre-defined task descriptions. These exploration methods are unaware of structures of the task distribution, thus are less effective in exploring coarser but useful information, i.e., clusters. Our cluster-aware exploration is designed to quickly reduce the uncertainty in cluster inference. As shown in Figure \ref{fig:milet_trace}, the agent explores on the map scale to find the most suitable structure.

To better understand the implication of our cluster-aware exploration, we compare the clustering quality (measured by NMI score) at the end of the first episode of \model{} and its variants without the exploration policy and without S-GRU in Table \ref{tab:nmi_explore}. Similar to VariBAD, the variant without the exploration policy performs exploration by considering task uncertainty implicitly. Our exploration policy with explicit clustering objective obtains better clustering quality, which builds foundation for refining task inference, resulting in better final performance in Table \ref{tab:ablation}. 

\begin{table}[!htp]
\centering
\caption{NMI score of different exploration methods.}
\label{tab:nmi_explore}
\begin{tabular}{c|c|c|c|c}
\hline
            & Ant-Goal & Humanoid-Dir & Hopper-Rand-Params & Walker-Rand-Params \\ \hline
$\neg$exploration & 0.466    & 0.485        & 0.517              & 0.327              \\ \hline
$\neg$S-GRU       &  0.374   &  0.512       &  0.552      &    0.468    \\ \hline
\model{}          & 0.615    & 0.639        & 0.588              & 0.765              \\ \hline
\end{tabular}
\end{table}

\end{document}